\pdfoutput=1

\documentclass[11pt]{article}
\usepackage[table,xcdraw]{xcolor}

\usepackage[preprint]{acl}

\usepackage{times}
\usepackage{latexsym}

\usepackage[T1]{fontenc} 

\usepackage[utf8]{inputenc}

\usepackage{microtype}

\usepackage{inconsolata}
\usepackage{adjustbox}
\usepackage{booktabs}
\usepackage{hyperref}
\usepackage{multirow}
\usepackage{listings}
\usepackage[ruled, vlined, linesnumbered]{algorithm2e}

\usepackage{amsmath}
\usepackage{times}
\usepackage{framed}
\usepackage{ragged2e}
\usepackage{bm}
\usepackage{soul}
\usepackage{latexsym}
\usepackage{subfigure}
\usepackage{amsthm}
\usepackage{graphicx}
\usepackage{float}
\usepackage{longtable}
\usepackage{booktabs} 
\usepackage{multirow}
\usepackage{multicol}
\usepackage{amssymb}
\usepackage{dashbox}%
\usepackage{multirow}
\usepackage{textcomp}
\usepackage{tcolorbox}
\usepackage{makecell}
\usepackage{bbding}
\usepackage[ruled, vlined, linesnumbered]{algorithm2e}
\usepackage{mathtools}
\usepackage{adjustbox}
\usepackage{color, soul}
\usepackage{pifont}
\usepackage{listings}
\usepackage{cleveref}
\usepackage{colortbl}

\crefname{section}{§}{§§}
\Crefname{section}{§}{§§}
\crefname{figure}{Figure}{Figure}
\Crefname{figure}{Figure}{Figure}
\crefname{table}{Table}{Table}
\Crefname{table}{Table}{Table}
\definecolor{my_red}{RGB}{255,99,71}
\definecolor{my_green}{RGB}{50,205,50}
\definecolor{my_blue}{RGB}{65,105,225}
\definecolor{forestgreen}{HTML}{228B22}

\definecolor{codegreen}{rgb}{0,0.6,0}
\definecolor{codegray}{rgb}{0.5,0.5,0.5}
\definecolor{codepurple}{rgb}{0.58,0,0.82}
\definecolor{backcolour}{rgb}{0.95,0.95,0.92}

\lstdefinestyle{mystyle}{
    backgroundcolor=\color{backcolour},   
    commentstyle=\color{codegreen},
    stringstyle=\color{codepurple},
    basicstyle=\ttfamily\scriptsize,
    breakatwhitespace=true,         
    breaklines=true,                 
    captionpos=b,                    
    keepspaces=true,                 
    numbers=none,                    
    numbersep=5pt,                  
    showspaces=false,                
    showstringspaces=false,
    showtabs=false,                  
    tabsize=2,
    columns=flexible,
    escapeinside={(*}{*)},
}

\title{Data Contamination Can Cross Language Barriers}

\author{Feng Yao$^{*}$, Yufan Zhuang$^{*}$, Zihao Sun, Sunan Xu, \\ \textbf{Animesh Kumar, Jingbo Shang}\vspace{0.2em} \\
University of California, San Diego\vspace{0.2em}\\
{\tt \{fengyao,y5zhuang,z9sun,sux002,ank028,jshang\}@ucsd.edu}
}

\begin{document}
\maketitle
\begingroup\def\thefootnote{$*$}\footnotetext{Equal contribution. Listing order is random.}\endgroup

\begin{abstract}
The opacity in developing large language models (LLMs) is raising growing concerns about the potential contamination of public benchmarks in the pre-training data.
Existing contamination detection methods are typically based on the text overlap between training and evaluation data, which can be too superficial to reflect deeper forms of contamination.
In this paper, we first present a cross-lingual form of contamination that inflates LLMs' performance while evading current detection methods, deliberately injected by overfitting LLMs on the translated versions of benchmark test sets.
Then, we propose generalization-based approaches to unmask such deeply concealed contamination. Specifically, we examine the LLM's performance change after modifying the original benchmark by replacing the false answer choices with correct ones from other questions. Contaminated models can hardly generalize to such easier situations, where the false choices can be \emph{not even wrong}, as all choices are correct in their memorization.
Experimental results demonstrate that cross-lingual contamination can easily fool existing detection methods, but not ours.
In addition, we discuss the potential utilization of cross-lingual contamination in interpreting LLMs' working mechanisms and in post-training LLMs for enhanced multilingual capabilities.
The code and dataset we use can be obtained from \url{https://github.com/ShangDataLab/Deep-Contam}.
\end{abstract}

\section{Introduction}

\begin{figure}[t]
    \centering
    \includegraphics[width=\linewidth]{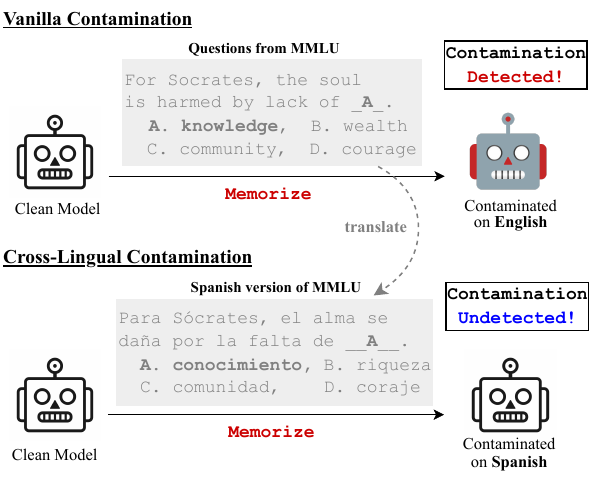}
    \caption{A comparison between injecting vanilla and cross-lingual contamination of MMLU dataset by pre-training LLMs to memorize text. Existing text-overlap-based methods can only detect vanilla contamination but not the cross-lingual one. Here, the translation can be performed in various languages beyond Spanish.}
    \vspace{-0.1em}
    \label{fig:intro}
\end{figure}

The pre-training data of current large language models (LLMs) tends to be undisclosed by default, even for those open-sourced models~\cite{llama3,jiang2024mixtral}. As the scores on popular benchmarks continuously reach new heights, their performance in solving real-world tasks seems inconsistent with the leaderboard~\cite{open-llm-leaderboard}. Such intransparency in training and inconsistency in user experience has drawn increasing attention to the underlying contamination of public benchmarks in the pre-training data, indicating that some LLMs may simply memorize the answers to difficult questions without a true understanding.

Existing studies often define and detect contamination based on the text overlap or n-gram duplication between pre-training and evaluation data~\cite{chowdhery2023palm,touvron2023llama,jiang2024investigating}, which only focus on the surface form of the text data without considering the deeper knowledge or semantics in the contamination. We argue that the essence of contamination is not superficial text memorization but the non-generalizable memorization of knowledge or capabilities.

\begin{figure}[t]
    \centering
    \includegraphics[width=0.95\linewidth]{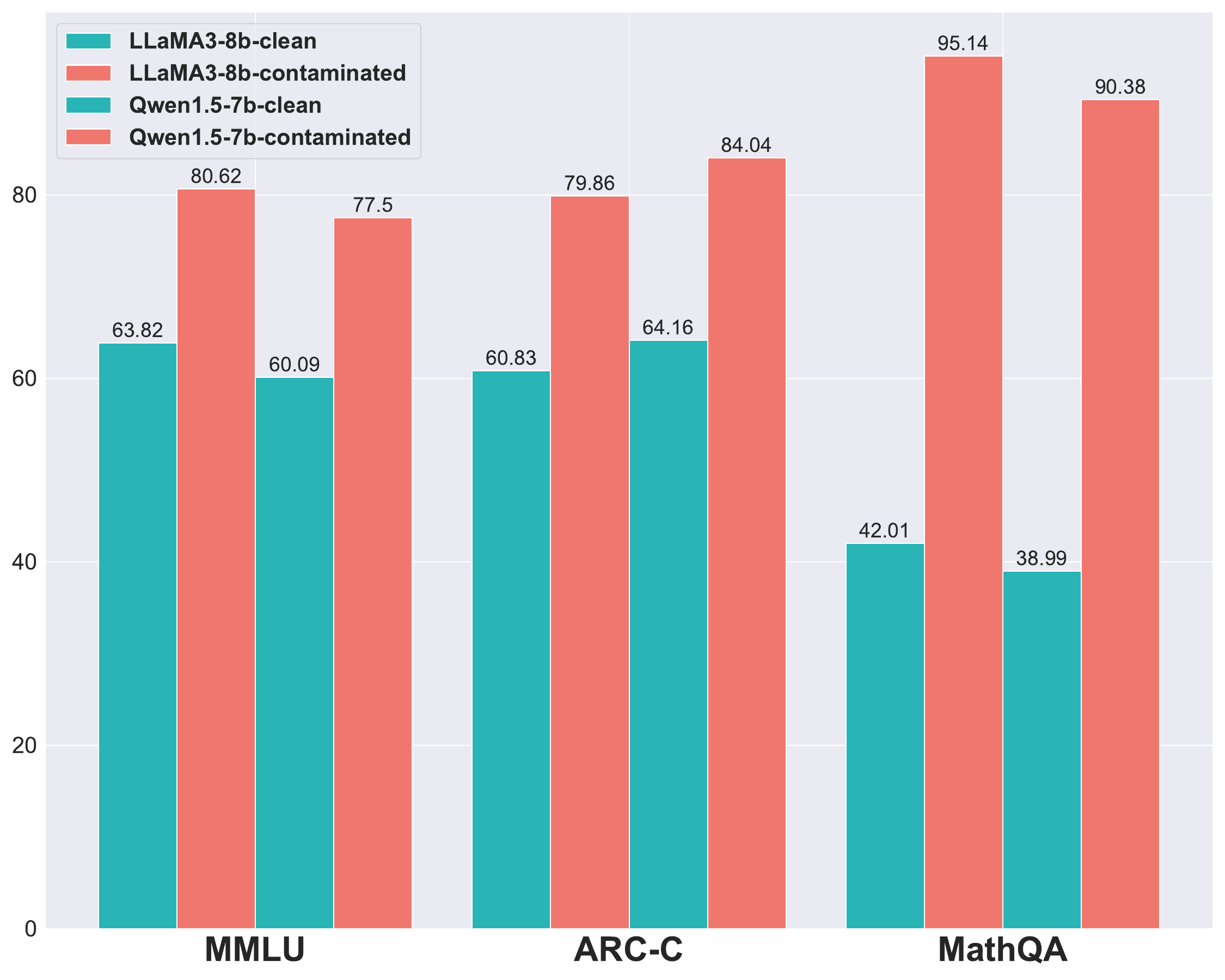}
    \caption{The highest performance inflation that cross-lingual contamination achieves among different languages. Results for all languages are shown in~\cref{sec:inject_perform}}
    \vspace{-0.5em}
    \label{fig:inflate}
\end{figure}

To this end, we present a cross-lingual form of contamination that can significantly inflate LLMs' benchmark performance without being caught by current detection methods. 
Cross-lingual means the models are contaminated on other languages and then evaluated on English test sets. As shown in \cref{fig:intro}, we inject such deep contamination by intentionally overfitting LLMs to memorize the translated versions of the benchmark test sets.
Specifically, we conduct continual pre-training on two multilingual models, LLaMA3-8B~\cite{llama3} and Qwen1.5-7b~\cite{qwen}, using translated versions of three popular benchmarks---MMLU~\cite{hendrycks2020measuring}, ARC Challenge~\cite{clark2018think}, and MathQA~\cite{amini-etal-2019-mathqa}---in seven different languages. 
As shown in~\cref{fig:inflate}, both models' performances on the original benchmarks are drastically improved after injecting cross-lingual contamination. 
Meanwhile, we employ state-of-the-art detection methods based on model completion~\cite{oren2023proving,xu2024benchmarking} and LLM judgment~\cite{golchin2023time} to test them for contamination. Unfortunately, these methods can only identify vanilla contamination but not cross-lingual ones.

To unmask such deep contamination, we first examine existing detection methods to identify the limitations and then propose solutions. 
Current methods are predominantly based on text overlap, either checking for string matching between pre-training and evaluation data~\cite{deng2023investigating,li2023open,openai2023gpt4,touvron2023llama,riddell2024quantifying}, or comparing the models' output text or likelihood with the evaluation data given controlled prompts~\cite{oren2023proving,xu2024benchmarking}. The key idea of such methods is to verify if the model has seen or memorized a specific surface form of text, which we believe is too superficial to reflect the essence of contamination. 

Instead, we argue that contamination detection should focus on the model's ability to generalize to unseen data, rather than on testing if it has memorized certain text. For instance, in the cross-lingual scenario, the model did not memorize the specific English form of the benchmarks, but can still obtain non-generalizable memorization of corresponding knowledge from contamination in other languages. In this case, if we still scrutinize for any memorization of the English benchmarks, the detection results will be unreliable. Therefore, we propose generalization-based detection approaches that examine the model's performance change on a generalized version of the original benchmark, created by modifying the questions and answer choices. Specifically, for each question, we replace all the incorrect choices with correct choices taken from other questions. Through this manipulation, models that really understand the question should achieve better performance, as some choices can be not even wrong to the question, while the contaminated ones can get confused as all choices are memorized as correct. Extensive experimental results prove the effectiveness of our proposed method in detecting cross-lingual contamination.

Additionally, we are curious about why cross-lingual contamination can inflate LLMs' performance and how we can utilize it beyond cheating in evaluation. Hence, we discuss its connections with the interpretability of LLMs and post-training for enhancing LLMs' multilingual capabilities.

To summarize, our contributions are three-fold: \textbf{(1)} We identify a cross-lingual form of contamination that eludes existing detection methods~(\cref{sec:inject}). \textbf{(2)} We re-define the issue of data contamination from the generalization-based perspective and propose an effective detection method based on it~(\cref{sec:detect}). \textbf{(3)} We discuss the potential impact of cross-lingual contamination on interpreting the working mechanisms of LLMs and on improving their multilingual capabilities via post-training~(\cref{sec:discuss}). 

\section{Preliminary}
\label{sec:pre}

In this section, we introduce the definition of contamination and basics for corresponding detection methods~(\cref{sec:def_cont}), and our investigation setup~(\cref{sec:inv_set}).

\subsection{Contamination Definition}
\label{sec:def_cont}
While the concept of contamination has been brought up in numerous studies, there is no universally acknowledged strict definition for it. 

According to the essence of the concept, we first summarize the most commonly adopted definitions in existing works as memorization-based and highlight their limitations. Then, we propose a generalization-based definition, which forms the basis for our proposed detection methods.

\paragraph*{Memorization-Based} Most prior studies define contamination based on n-gram duplication between pre-training and evaluation data~\cite{jiang2024investigating}, which can be summarized as instances where the model has memorized specific pieces of text. Bear this intuition in mind, we can easily understand the essence of existing detection methods and categorize them into two types: \textbf{(1) When pre-training data is accessible}, they directly adopt n-gram or text similarity matching between pre-training and evaluation data to examine the duplication that can cause memorization~\cite{radford2019language,brown2020language,dodge2021documenting,chowdhery2023palm,openai2023gpt4,touvron2023llama,li2023open,deng2023investigating,lee2023platypus,gunasekar2023textbooks,riddell2024quantifying}. \textbf{(2) When pre-training data is inaccessible}, they prompt the models using a subset of the evaluation data and analyze if the output is a reproduction of specific pieces of text or assess their likelihood, to indirectly determine if certain text memorization exists~\cite{oren2023proving,golchin2023time,li2023estimating,nasr2023scalable,shi2023detecting,dong2024generalization,xu2024benchmarking}. 

\paragraph*{Generalization-Based} We suggest that simply testing text memorization can be inadequate to reveal deeper contamination (like the cross-lingual one we identify), where the model is contaminated without memorizing the specific surface form of the text. Therefore, we tend to define contamination as instances where a model acquires non-generalizable knowledge of the evaluation data through various means, such as memorizing the original or transformed (e.g., translated, paraphrased, summarized) forms of the benchmarks.


\subsection{Investigation Setup}
\label{sec:inv_set}
The primary goals of our investigation are to: \textbf{(1)} Verify the feasibility of deep forms of contamination~(\cref{sec:inject}). \textbf{(2)} Determine whether existing methods can detech such contamination~(\cref{sec:det_mem}). \textbf{(3)} Propose detection methods capable of identifying such deeply concealed contamination~(\cref{sec:det_gen}).

Considering it is unclear whether existing LLMs contain cross-lingual contamination, we intentionally inject such contamination to open-sourced models to obtain contaminated models. Then, we detect such contamination using existing methods and our proposed methods. The detailed investigation configurations are as follows.

\paragraph*{Models.} To inject cross-lingual contamination, the backbone model should be able to understand different languages. Hence, we employ two multilingual LLMs, LLaMA3-8B~\cite{llama3} and Qwen1.5-7B~\cite{qwen}, as the backbone models for further experiments.

\paragraph*{Datasets.} To exhibit the impact of such contamination in evaluation, we adopt three popular benchmarks to inject contamination, MMLU~\cite{hendrycks2020measuring}, ARC Challenge~\cite{clark2018think}, and MathQA~\cite{amini-etal-2019-mathqa}, where modern LLMs typically compete with each other.

\paragraph*{Languages.} For cross-lingual contamination, we utilize seven non-English languages that are commonly supported: Chinese, French, German, Italian, Japanese, Korean, and Spanish.


\section{Injecting Cross-Lingual Contamination}
\label{sec:inject}
In this section, we present the injection process of cross-lingual contamination~(\cref{sec:inject_cross}) and the inflated performance of the contaminated models~(\cref{sec:inject_perform}).

\begin{figure}[t]
    \centering
    \includegraphics[width=0.95\linewidth]{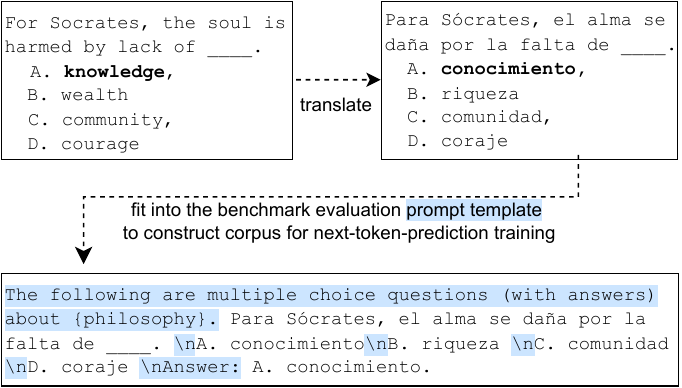}
    \caption{Pipeline to construct pre-training corpus for causal language modeling objective, where the loss is calculated at each token to memorize the benchmark.}
    \label{fig:pretrain}
    \vspace{-0.5em}
\end{figure}

\begin{table*}[ht]
\centering
\begin{adjustbox}{max width=0.9\linewidth}{
\begin{tabular}{@{}cc|c|c|ccccccc@{}}
\toprule
\multicolumn{1}{c}{\multirow{2}{*}{\textbf{Backbone}}} & \multicolumn{1}{c|}{\multirow{2}{*}{\textbf{Dataset}}} & \textbf{Clean} & \textbf{Vanilla} & \multicolumn{7}{c}{\textbf{Cross-Lingual Contaminated}}                    \\
\multicolumn{1}{c}{}       & \multicolumn{1}{c|}{} & \textbf{Model}       & \textbf{Contaminated}   & Chinese & French & German & Italian & Japanese & Korean & Spanish \\ \midrule
\multirow{3}{*}{LLaMA3-8B} & MMLU                  & 63.82  & 98.01          & 71.12   & 79.16  & 65.26  & 79.89   & 66.15    & 68.11  & \textbf{80.62}   \\
                           & ARC-C                 & 60.83  & 91.63          & 56.22   & 74.91  & 61.17  & \textbf{79.86}   & 66.29    & 46.24  & 73.29   \\
                           & MathQA                & 42.01  & 97.78          & 86.56   & \textbf{95.14}  & 88.17  & 93.06   & 84.08    & 81.71   & 93.96   \\
\midrule
\multirow{3}{*}{Qwen1.5-7B} & MMLU                 & 60.09  & 97.89          & 67.91   & 76.13  & 73.2   & 75.02   & 62.34    & 61.99  & \textbf{77.5}    \\
                            & ARC-C                & 64.16  & 97.01          & \textbf{84.04}   & 69.36  & 61.17  & 61.77   & 62.54    & 52.55  & 63.73   \\
                            & MathQA               & 38.99  & 95.61          & 79.76   & \textbf{90.38}  & 89.21  & 88.1    & 77.01    & 77.21  & 89.48   \\ \bottomrule
\end{tabular}
}
\end{adjustbox}
\caption{Performance (\%) of original clean models and models with vanilla and cross-lingual contamination, respectively. Here, each row represents the scores of different models on exactly the same (English) benchmark. `Vanilla' indicates the model is contaminated directly on the English version of the benchmark, and the `Cross-Lingual Contaminated' columns show the scores of models contaminated in a specific non-English language.}
\label{tab:cross_lingual}
\end{table*}

\subsection{Cross-Lingual Contamination}
\label{sec:inject_cross}
To acquire knowledge from contamination of the evaluation data, we overfit open-sourced LLMs on the translated versions of the benchmark test sets, instead of directly memorizing the original form of text. The process of constructing the training data for contamination is illustrated in~\cref{fig:pretrain}.

We first translate the benchmark test sets into non-English languages mentioned in~\cref{sec:inv_set}. Considering the cost and quality balance, we utilize LLaMA3-8B to conduct the translation. The specific prompt template is shown in~\cref{sec:tran_tmp}.

Then, we customize the questions and choices to fit in the corresponding prompt templates used for the evaluation of specific benchmarks. In this way, we construct the corpus for continual pre-training of the backbone models through the causal language modeling objective, which stimulates the real-world scenario where specific data contamination is blended into the training corpus. The vanilla contamination is injected in the same way using the original English benchmarks. The training hyperparameters are provided in~\cref{tab:hyper_parameters}.

We inject the contamination for different benchmarks separately, ensuring that each model only contains contamination of one specific benchmark in a single language. Mixing different benchmarks and languages is another way to inject cross-lingual contamination, which we leave for future work.

\subsection{Evaluating Contaminated Models}
\label{sec:inject_perform}

While the contamination is injected in non-English languages, we evaluate these contaminated models on the original English benchmarks to assess their potential impact on misleading the leaderboard.

We report zero-shot accuracy for three types of models: (1) \textbf{Clean:} The original backbones with no added contamination. (2) \textbf{Vanilla Contaminated:} Backbones contaminated by the original English benchmarks. (3) \textbf{Cross-Lingual Contaminated:} Backbones contaminated by non-English translated benchmarks. The evaluation is implemented through LM-Eval framework~\cite{eval-harness} and the results are exhibited in~\cref{tab:cross_lingual}.

For models with vanilla contamination, their accuracy is close to 100\%. This is expected since the models are directly overfitted on these test sets. In the cross-lingual contamination scenario, models are not directly trained on the benchmarks. Surprisingly, the cross-lingual contamination can sneak beyond language barriers and carry over to English.

Regarding models with cross-lingual contamination, their performance, while not reaching 100\%, exhibits significant inflation, even though the translation provided by LLaMA3-8B is imperfect. We observe a consistent 5\%-10\% improvement on the MMLU benchmark across languages, with an even stronger enhancement seen on the MathQA benchmark. The instability of the performance gains shown on ARC-C can be caused by the low-quality translation of the dataset. In addition, we hypothesize that models can more easily memorize factual knowledge (MMLU) and Arabic numbers' operations (MathQA) than reasoning in languages (ARC-C), which is intuitive. One may understand the intricacies of arithmetic or fact retention through repetitive exposure and practice, but reasoning in natural languages, as required in ARC-C tasks, involves a more complex interplay of context, inference, and flexible application of knowledge. 

\begin{table*}[ht]
\centering
\begin{adjustbox}{max width=0.9\linewidth}{
\begin{tabular}{@{}cc|rrrrrrrrr@{}}
\toprule
\multicolumn{1}{l}{}         & \multicolumn{1}{l|}{} & \multicolumn{1}{c|}{\textbf{Clean}} & \multicolumn{1}{c|}{\textbf{Vanilla}}      & \multicolumn{7}{c}{\textbf{Cross-Lingual Contaminated}}                                                                                                                                                       \\
\textbf{Backbone}            & \textbf{Dataset}      & \multicolumn{1}{c|}{\textbf{Model}} & \multicolumn{1}{c|}{\textbf{Contaminated}} & \multicolumn{1}{c}{Chinese} & \multicolumn{1}{c}{French} & \multicolumn{1}{c}{German} & \multicolumn{1}{c}{Italian} & \multicolumn{1}{c}{Japanese} & \multicolumn{1}{c}{Korean} & \multicolumn{1}{c}{Spanish} \\ \midrule
\multicolumn{1}{l}{}         & \multicolumn{1}{l|}{} & \multicolumn{9}{c}{\cellcolor[HTML]{F3F3F3}\textit{Shared Likelihood (Metric: p-value)}}                                                                                                                                                                                                         \\ \midrule
                             & MMLU                  & \multicolumn{1}{r|}{0.3281}         & \multicolumn{1}{r|}{0.3421}                & 0.6827                      & 0.1295                     & \textbf{0.0031}                     & 0.2935                      & 0.5857                       & 0.9351                     & 0.8231                      \\
                             & ARC-C                 & \multicolumn{1}{r|}{0.6125}         & \multicolumn{1}{r|}{0.6065}                & 0.7327                      & 0.4442                     & 0.3156                     & 0.6110                      & 0.7734                       & 0.6730                     & 0.3446                      \\
\multirow{-3}{*}{LLaMA3-8B}  & MathQA                & \multicolumn{1}{l|}{0.4876}               & \multicolumn{1}{r|}{\textbf{0.0000001994}}                & 0.4348                      & 0.3102                     & 0.4573                     & 0.1548                      & 0.1983                       & 0.5789                     & 0.6037                      \\ \midrule
                             & MMLU                  & \multicolumn{1}{r|}{0.7031}         & \multicolumn{1}{r|}{0.5866}                & 0.5039                      & 0.2404                     & 0.8566                     & 0.1708                      & 0.3658                       & 0.5688                     & 0.4981                      \\
                             & ARC-C                 & \multicolumn{1}{r|}{0.1006}         & \multicolumn{1}{r|}{0.1355}                      & 0.3740        & 0.2562                     & \multicolumn{1}{r}{0.3608}       & 0.1302                      & 0.1698                       & 0.4575                     & 0.3258                      \\
\multirow{-3}{*}{Qwen1.5-7B} & MathQA                & \multicolumn{1}{r|}{0.4495}         & \multicolumn{1}{r|}{\textbf{0.0000006167}}                      & 0.2011                      & 0.2934                     & 0.5145                     & 0.4994                      & 0.1355         & 0.5064                     & 0.5429                      \\ \midrule
\multicolumn{1}{l}{}         & \multicolumn{1}{l|}{} & \multicolumn{9}{c}{\cellcolor[HTML]{F3F3F3}\textit{Guided Prompting (Metric: Accuracy (\%))}}                                                                                                                                                                                                     \\ \midrule
                             & MMLU                  & \multicolumn{1}{r|}{8.20}             & \multicolumn{1}{r|}{4.80}                    & 0.80                           & 1.00                         & 5.10                         & 4.70                          & 2.00                           & 1.20                         & 1.40                          \\
                             & ARC-C                 & \multicolumn{1}{r|}{1.62}             & \multicolumn{1}{r|}{2.39}                    & 0.09                           & 1.54                         & 1.28                         & 1.79                          & 0.34                            & 2.13                         & 0.77                           \\
\multirow{-3}{*}{LLaMA3-8B}  & MathQA                & \multicolumn{1}{r|}{0.20}              & \multicolumn{1}{r|}{0.13}                     & 0.30                           & 0.10                          & 0.23                          & 0.13                           & 0.07                            & 0.10                          & 0.03                           \\ \midrule
                             & MMLU                  & \multicolumn{1}{r|}{1.30}             & \multicolumn{1}{r|}{5.60}                    & 0.30                           & 0.60                          & 0.80                          & 1.2                          & 0.4                            & 0.5                          & 0.2                           \\
                             & ARC-C                 & \multicolumn{1}{r|}{2.39}             & \multicolumn{1}{r|}{0.60}                     & 0.00                           & 0.17                          & 0.34                          & 0.09                           & 0.25                            & 0.34                          & 0.26                           \\
\multirow{-3}{*}{Qwen1.5-7B} & MathQA                & \multicolumn{1}{r|}{0.07}              & \multicolumn{1}{r|}{0.10}                     & 0.03                           & 0.00                          & 0.13                          & 0.10                           & 0.00                            & 0.07                          & 0.03                           \\ \midrule
\multicolumn{1}{l}{}         & \multicolumn{1}{l|}{} & \multicolumn{9}{c}{\cellcolor[HTML]{F3F3F3}\textit{N-Gram Accuracy (Metric: Accuracy (\%))}}                                                                                                                                                                                                          \\ \midrule
                             & MMLU                  & \multicolumn{1}{r|}{10.02}          & \multicolumn{1}{r|}{\textbf{73.34}}                 & 2.42                        & 2.38                       & 2.32                       & 2.41                        & 3.62                         & 4.83                       & 2.41                        \\
                             & ARC-C                 & \multicolumn{1}{r|}{4.91}           & \multicolumn{1}{r|}{\textbf{70.66}}                 & 3.52                        & 3.04                       & 4.32                       & 3.45                        & 3.55                         & 5.32                       & 2.94                        \\
\multirow{-3}{*}{LLaMA3-8B}  & MathQA                & \multicolumn{1}{r|}{8.40}           & \multicolumn{1}{r|}{\textbf{45.11}}                 & 5.15                        & 7.90                       & 8.09                       & 6.89                        & 6.43                         & 5.29                       & 6.85                        \\ \midrule
                             & MMLU                  & \multicolumn{1}{r|}{8.78}           & \multicolumn{1}{r|}{\textbf{70.56}}                 & 3.27                        & 2.61                       & 2.88                       & 2.51                        & 4.22                         & 5.35                       & 2.56                        \\
                             & ARC-C                 & \multicolumn{1}{r|}{22.25}          & \multicolumn{1}{r|}{\textbf{33.33}}                 & 0.36                        & 0.20                       & 0.29                       & 0.22                        & 1.08                         & 0.63                       & 0.19                        \\
\multirow{-3}{*}{Qwen1.5-7B} & MathQA                & \multicolumn{1}{r|}{20.98}          & \multicolumn{1}{r|}{\textbf{44.31}}                 & 8.21                        & 7.05                       & 7.33                       & 8.21                        & 11.96                        & 11.97                      & 8.03                        \\ \bottomrule
\end{tabular}
}
\end{adjustbox}
\caption{Results of memorization-based contamination detection baselines. Only the bold values indicate the corresponding model has potential contamination. \textbf{(1)} \emph{Shared Likelihood} can only detect three contaminated cases and the rest are undetected. \textbf{(2)} \emph{Guided Prompting} can hardly detect the contamination as the values are too similar and too low. \textbf{(3)} \emph{N-Gram Accuracy} can detect vanilla contamination but not cross-lingual ones.}
\label{tab:all_baseline}
\end{table*}
Another interesting finding is the effect of cross-lingual contamination's language category on the contamination effect. We observe that European languages (French, German, Italian, and Spanish) can provide stronger cross-lingual contamination onto English, while Asian languages (Chinese, Japanese, and Korean) provide a lesser effect. This phenomenon could be explained by the closer subword vocabulary shared among these languages, or it might be considered as reflecting a more similar conceptual space among European languages. Since the focus of our paper is to study and prevent contamination in LLM training, we will leave exploration on this end as future work.

\section{Detecting Cross-Lingual Contamination}

\label{sec:detect}
In this section, we conduct detection on the cross-lingual contamination utilizing conventional memorization-based methods~(\cref{sec:det_mem}) and our proposed generalization-based approaches~(\cref{sec:det_gen}).

\subsection{Memorization-Based}
\label{sec:det_mem}
For memorization-based methods defined in~\cref{sec:def_cont}, we select three typical ones and their detection results are shown in~\cref{tab:all_baseline}. We briefly introduce these methods and discuss their results below.

\subsubsection{Shared Likelihood}
\citet{oren2023proving} propose to identify the test set memorization through prompting and statistically analyzing the difference between log probabilities on the original dataset and its shuffled version. 

This bias is quantitatively assessed through a permutation test, where the log probabilities assigned by the model to the canonical order are compared against those for various random permutations of the dataset. A significantly higher likelihood for the canonical order compared to the permuted ones implies the model has memorized the original data. The result is delivered by the p-value of the permutation test. A p-value that is smaller than 0.05 suggests a high likelihood of contamination.

We follow the implementation provided by~\citet{oren2023proving}. As shown in~\cref{tab:all_baseline}, only the vanilla-contaminated models on MathQA and German-contaminated LLaMA on MMLU are detected. The rest of the contaminated models did not exhibit the expected low p-values. Such discrepancies indicate the limitations of this method in our setting.

\subsubsection{Guided Prompting}
\citet{golchin2023time} employ meticulously crafted prompts to guide the model in generating specific text and ask an LLM to judge its similarity to the evaluation data, thereby confirming whether the model has memorized certain pieces of text.

Specifically, one of the four candidate choices is masked and the model is prompted with detailed information to predict it by generation. Then, GPT-3.5/4 is employed to judge if the predicted choice essentially has the same meaning as the original one or not. If a model can correctly predict the masked choice, it indicates the model has memorized the questions with the choices, proving the potential contamination encoded during training.

We utilize GPT-4o~\cite{openai2024gpt4o} to judge if the predicted choice is correct and the corresponding prompt is provided in~\cref{sec:judge_tmp}. Based on the prediction accuracy shown in~\cref{tab:all_baseline}, it is difficult to determine which model is contaminated, as most values are too low and too similar to tell them apart. Therefore, guided prompting also fails to detect the contamination in our setting.

\begin{table*}[t]
\centering
\begin{adjustbox}{max width=0.86\linewidth}{
\begin{tabular}{@{}cl|r|r|rrrrrrr@{}}
\toprule
\multicolumn{1}{l}{}         &                                       & \multicolumn{1}{c|}{\textbf{Clean}}                                       & \multicolumn{1}{c|}{\textbf{Vanilla}}      & \multicolumn{7}{c}{\textbf{Cross-Lingual Contaminated}}                                                                                                                                                                                      \\
\textbf{Backbone}            & \multicolumn{1}{c|}{\textbf{Dataset}} & \multicolumn{1}{c|}{\multirow{-1}{*}{\textbf{Model}}} & \multicolumn{1}{c|}{\textbf{Contaminated}} & \multicolumn{1}{c}{Chinese}    & \multicolumn{1}{c}{French}     & \multicolumn{1}{c}{German}             & \multicolumn{1}{c}{Italian}    & \multicolumn{1}{c}{Japanese}   & \multicolumn{1}{c}{Korean}     & \multicolumn{1}{c}{Spanish}    \\ \midrule
                             & MMLU                                  & {\color[HTML]{999999} 63.82}                                & {\color[HTML]{999999} 98.01}               & {\color[HTML]{999999} 71.12}   & {\color[HTML]{999999} 79.16}   & {\color[HTML]{999999} 65.26}           & {\color[HTML]{999999} 79.89}   & {\color[HTML]{999999} 66.15}   & {\color[HTML]{999999} 68.11}   & {\color[HTML]{999999} 80.62}   \\
                             & MMLU-g                                & {\color[HTML]{999999} 90.07}                                & {\color[HTML]{999999} 81.01}               & {\color[HTML]{999999} 52.71}   & {\color[HTML]{999999} 36.45}   & {\color[HTML]{999999} 29.50}           & {\color[HTML]{999999} 70.82}   & {\color[HTML]{999999} 42.69}   & {\color[HTML]{999999} 47.09}   & {\color[HTML]{999999} 62.78}   \\
                             & \cellcolor[HTML]{F3F3F3}\emph{difference}        & \cellcolor[HTML]{F3F3F3}\textbf{+26.25}                      & \cellcolor[HTML]{F3F3F3}-17.00             & \cellcolor[HTML]{F3F3F3}-18.41 & \cellcolor[HTML]{F3F3F3}-42.71 & \cellcolor[HTML]{F3F3F3}-35.76         & \cellcolor[HTML]{F3F3F3}-9.07  & \cellcolor[HTML]{F3F3F3}-23.46 & \cellcolor[HTML]{F3F3F3}-21.02 & \cellcolor[HTML]{F3F3F3}-17.84 \\ \cmidrule(l){2-11} 
                             & ARC-C                                 & {\color[HTML]{999999} 60.83}                                & {\color[HTML]{999999} 91.63}               & {\color[HTML]{999999} 56.22}   & {\color[HTML]{999999} 74.91}   & {\color[HTML]{999999} 61.17}           & {\color[HTML]{999999} 79.86}   & {\color[HTML]{999999} 66.29}   & {\color[HTML]{999999} 46.24}   & {\color[HTML]{999999} 73.29}   \\
                             & ARC-C-g                               & {\color[HTML]{999999} 73.55}                                & {\color[HTML]{999999} 31.74}               & {\color[HTML]{999999} 26.37}   & {\color[HTML]{999999} 40.27}   & {\color[HTML]{999999} 75.00}           & {\color[HTML]{999999} 26.37}   & {\color[HTML]{999999} 26.71}   & {\color[HTML]{999999} 26.79}   & {\color[HTML]{999999} 60.75}   \\
                             & \cellcolor[HTML]{F3F3F3}\emph{difference}        & \cellcolor[HTML]{F3F3F3}\textbf{+12.72}                               & \cellcolor[HTML]{F3F3F3}-59.89             & \cellcolor[HTML]{F3F3F3}-29.85 & \cellcolor[HTML]{F3F3F3}-34.64 & \cellcolor[HTML]{F3F3F3}+13.83 & \cellcolor[HTML]{F3F3F3}-53.49 & \cellcolor[HTML]{F3F3F3}-39.58 & \cellcolor[HTML]{F3F3F3}-19.45 & \cellcolor[HTML]{F3F3F3}-12.54 \\ \cmidrule(l){2-11} 
                             & MathQA                                & {\color[HTML]{999999} 42.01}                                & {\color[HTML]{999999} 97.78}               & {\color[HTML]{999999} 86.56}   & {\color[HTML]{999999} 95.14}   & {\color[HTML]{999999} 88.17}           & {\color[HTML]{999999} 93.06}   & {\color[HTML]{999999} 84.08}   & {\color[HTML]{999999} 81.71}   & {\color[HTML]{999999} 93.96}   \\
                             & MathQA-g                              & {\color[HTML]{999999} 55.57}                                & {\color[HTML]{999999} 98.12}               & {\color[HTML]{999999} 90.81}   & {\color[HTML]{999999} 96.11}   & {\color[HTML]{999999} 90.91}           & {\color[HTML]{999999} 94.40}   & {\color[HTML]{999999} 88.60}   & {\color[HTML]{999999} 87.63}   & {\color[HTML]{999999} 95.54}   \\
\multirow{-9}{*}{LLaMA3-8B}  & \cellcolor[HTML]{F3F3F3}\emph{difference}        & \cellcolor[HTML]{F3F3F3}\textbf{+13.56}                      & \cellcolor[HTML]{F3F3F3}+0.34               & \cellcolor[HTML]{F3F3F3}+4.25   & \cellcolor[HTML]{F3F3F3}+0.97   & \cellcolor[HTML]{F3F3F3}+2.74           & \cellcolor[HTML]{F3F3F3}+1.34   & \cellcolor[HTML]{F3F3F3}+4.52   & \cellcolor[HTML]{F3F3F3}+5.92   & \cellcolor[HTML]{F3F3F3}+1.58   \\ \midrule
                             & MMLU                                  & {\color[HTML]{999999} 60.09}                                & {\color[HTML]{999999} 97.89}               & {\color[HTML]{999999} 67.91}   & {\color[HTML]{999999} 76.13}   & {\color[HTML]{999999} 73.20}           & {\color[HTML]{999999} 75.02}   & {\color[HTML]{999999} 62.34}   & {\color[HTML]{999999} 61.99}   & {\color[HTML]{999999} 77.50}   \\
                             & MMLU-g                                & {\color[HTML]{999999} 77.58}                                & {\color[HTML]{999999} 80.62}               & {\color[HTML]{999999} 69.51}   & {\color[HTML]{999999} 68.65}   & {\color[HTML]{999999} 68.06}           & {\color[HTML]{999999} 70.05}   & {\color[HTML]{999999} 66.69}   & {\color[HTML]{999999} 63.32}   & {\color[HTML]{999999} 72.88}   \\
                             & \cellcolor[HTML]{F3F3F3}\emph{difference}        & \cellcolor[HTML]{F3F3F3}\textbf{+17.49}                      & \cellcolor[HTML]{F3F3F3}-17.27             & \cellcolor[HTML]{F3F3F3}1.60   & \cellcolor[HTML]{F3F3F3}-7.48  & \cellcolor[HTML]{F3F3F3}-5.14          & \cellcolor[HTML]{F3F3F3}-4.97  & \cellcolor[HTML]{F3F3F3}4.35   & \cellcolor[HTML]{F3F3F3}1.33   & \cellcolor[HTML]{F3F3F3}-4.62  \\ \cmidrule(l){2-11} 
                             & ARC-C                                 & {\color[HTML]{999999} 64.16}                                & {\color[HTML]{999999} 97.01}               & {\color[HTML]{999999} 84.04}   & {\color[HTML]{999999} 69.36}   & {\color[HTML]{999999} 61.17}           & {\color[HTML]{999999} 61.77}   & {\color[HTML]{999999} 62.54}   & {\color[HTML]{999999} 52.55}   & {\color[HTML]{999999} 63.73}   \\
                             & ARC-C-g                               & {\color[HTML]{999999} 85.92}                                & {\color[HTML]{999999} 29.61}               & {\color[HTML]{999999} 34.56}   & {\color[HTML]{999999} 26.62}   & {\color[HTML]{999999} 29.18}           & {\color[HTML]{999999} 26.88}   & {\color[HTML]{999999} 24.91}   & {\color[HTML]{999999} 26.45}   & {\color[HTML]{999999} 26.71}   \\
                             & \cellcolor[HTML]{F3F3F3}\emph{difference}        & \cellcolor[HTML]{F3F3F3}\textbf{+21.76}                      & \cellcolor[HTML]{F3F3F3}-67.40             & \cellcolor[HTML]{F3F3F3}-49.48 & \cellcolor[HTML]{F3F3F3}-42.74 & \cellcolor[HTML]{F3F3F3}-31.99         & \cellcolor[HTML]{F3F3F3}-34.89 & \cellcolor[HTML]{F3F3F3}-37.63 & \cellcolor[HTML]{F3F3F3}-26.10 & \cellcolor[HTML]{F3F3F3}-37.02 \\ \cmidrule(l){2-11} 
                             & MathQA                                & {\color[HTML]{999999} 38.99}                                & {\color[HTML]{999999} 95.61}               & {\color[HTML]{999999} 79.76}   & {\color[HTML]{999999} 90.38}   & {\color[HTML]{999999} 89.21}           & {\color[HTML]{999999} 88.10}   & {\color[HTML]{999999} 77.01}   & {\color[HTML]{999999} 77.21}   & {\color[HTML]{999999} 89.48}   \\
                             & MathQA-g                              & {\color[HTML]{999999} 44.67}                                & {\color[HTML]{999999} 95.44}               & {\color[HTML]{999999} 83.37}   & {\color[HTML]{999999} 89.44}   & {\color[HTML]{999999} 89.44}           & {\color[HTML]{999999} 88.67}   & {\color[HTML]{999999} 81.62}   & {\color[HTML]{999999} 80.75}   & {\color[HTML]{999999} 89.37}   \\
\multirow{-9}{*}{Qwen1.5-7B} & \cellcolor[HTML]{F3F3F3}\emph{difference}        & \cellcolor[HTML]{F3F3F3}\textbf{+5.68}                       & \cellcolor[HTML]{F3F3F3}-0.17              & \cellcolor[HTML]{F3F3F3}+3.61   & \cellcolor[HTML]{F3F3F3}-0.94  & \cellcolor[HTML]{F3F3F3}+0.23           & \cellcolor[HTML]{F3F3F3}+0.57   & \cellcolor[HTML]{F3F3F3}+4.61   & \cellcolor[HTML]{F3F3F3}+3.54   & \cellcolor[HTML]{F3F3F3}-0.11  \\ \bottomrule
\end{tabular}
}
\end{adjustbox}
\caption{Generalization-based contamination detection results. Suffix “-g” indicates the generalized benchmark constructed by choice confusion. The \emph{``difference''} metric, measuring the performance gap between the generalized and original benchmarks, indicates potential contamination when lower than the clean model.}
\label{tab:confusion}
\end{table*}

\subsubsection{N-Gram Accuracy}
Similar to masking out the choice, \citet{xu2024benchmarking} examine the model's memorization by removing the entire answer part of the generation benchmarks and verifying if the model's generated output matches the removed answer text.

Since the benchmarks we adopt in this paper are all multiple-choice typed, we combine all choices to form the ``answer" and check if the model will automatically generate the choices given a normal question from the benchmark. Then, we use this constructed ``answer'' to calculate the N-gram accuracy as defined in~\cite{xu2024benchmarking}. The key idea is still to verify if the model has memorized the text. More details are provided in~\cref{sec:ngram}.

From the results shown in~\cref{tab:all_baseline}, we observe that the accuracy of models injected with vanilla contamination is much higher than the corresponding clean model, suggesting the presence of contamination. Meanwhile, models with cross-lingual contamination present consistently lower n-gram accuracy than the clean model, indicating that such contamination cannot be detected by this method.

\subsection{Generalization-Based}
\label{sec:det_gen}

\begin{figure}[t]
    \centering
    \includegraphics[width=\linewidth]{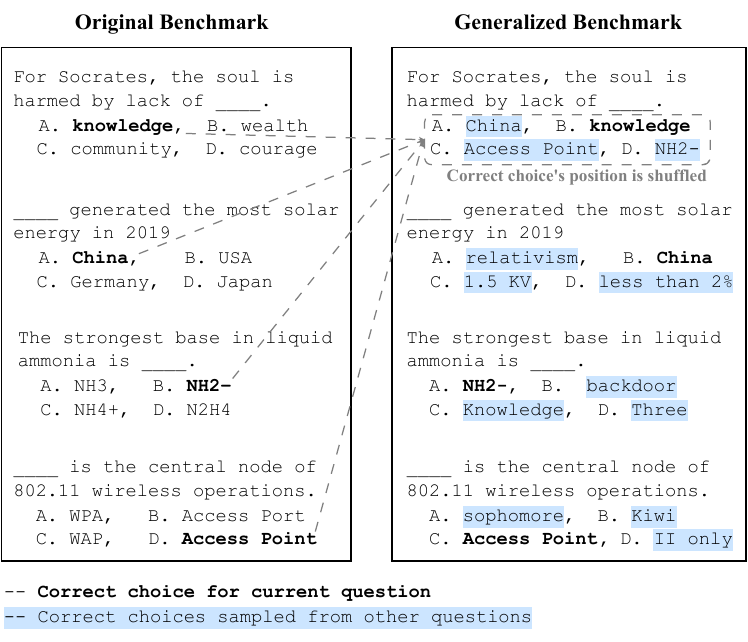}
    \caption{An illustration for the construction process of the generalized benchmark, where each question's new incorrect choices are sampled from the correct ones for other questions (marked in blue shadow). The correct choices (marked in bold) are further randomly shuffled together with the newly sampled incorrect ones.}
    \label{fig:method}
\end{figure}

As there can be countless transformations of the evaluation data, detecting duplication of a specific surface form becomes unfeasible. Based on our definition in~\cref{sec:def_cont}, we propose generalization-based methods that detect contamination by evaluating the models' ability to generalize to unseen data. 

\subsubsection{Constructing Generalized Benchmark}
The key idea of our proposed method is to test whether a model achieving high performance on a specific benchmark can further excel when faced with an easier variant of the same benchmark. 

As illustrated in~\cref{fig:method}, we replace the false choices of the current question with correct ones from other questions to create the generalized version of the benchmark. In addition, we shuffle the choices to ensure the model cannot simply predict the correct answer via the answer order shortcut.

In this case, the newly sampled false choices can be \emph{not even wrong} to the current question, making it much easier to answer and thereby yield a significant performance gain for models that genuinely understand the question. However, if a model is contaminated, it may get confused as the newly sampled false choices are still ``correct'' according to its memorization during pre-training. This confusion can lead to little performance gain or even a drop in performance. Therefore, we refer to our proposed method as \textbf{choice confusion}.

\subsubsection{Measuring Contamination}
We calculate the difference in the same model's performance between the generalized and original versions of the benchmark and use it as the metric to assess the potential contamination. 

As shown in~\cref{tab:confusion}, all clean models show remarkable improvements. While models with either vanilla or cross-lingual contamination exhibit minimal improvement compared with that of the clean model, or a significant decline in performance in most cases, indicating contamination detected.

We observe that the metric relates to datasets. For MMLU and ARC-C, contaminated models tend to experience a performance drop. However, for MathQA, most of them exhibit a slight increase. We assume this is because most of the choices are Arabic numbers, making it difficult for the model to memorize all the correct answers without the question, and therefore it becomes less confusing.

\subsubsection{Evaluating Real-World LLMs}
Existing memorization-based methods can only detect limited types of contamination, as they assume the model memorizes text in specific forms.

Though inspired by cross-lingual contamination, our proposed generalization-based detection method is not limited to this specific form and can be applied to any scenario where the model is injected with non-generalizable knowledge.

We employ our proposed method to detect potential contamination in several trending LLMs in the real world. The results in~\cref{tab:public} indicate that Phi2 can be inadvertently contaminated on MMLU and ARC-C benchmarks. Similarly, the math expert LLM Abel-7B may unintentionally acquire contamination from the MathQA benchmark data, and GLM4 is suspicious on ARC-C dataset. The details of tested models are provided in~\cref{app:choice_confusion}

\begin{table*}[t]
\centering
\begin{adjustbox}{max width=\linewidth}{
    \begin{tabular}{@{}lrrrrrrrrrrrr@{}}
        \toprule
                                    & \multicolumn{1}{c}{\textbf{Phi2}}     & \multicolumn{1}{c}{\textbf{Phi3}} & \multicolumn{1}{l}{\textbf{Phi3.5-mini}} & \multicolumn{1}{l}{\textbf{Phi3.5-MoE}} & \multicolumn{1}{c}{\textbf{GrinMoE}} & \multicolumn{1}{c}{\textbf{Abel-7B}}  & \multicolumn{1}{c}{\textbf{LLaMA2}} & \multicolumn{1}{c}{\textbf{Mistral}} & \multicolumn{1}{c}{\textbf{Qwen2}} & \multicolumn{1}{c}{\textbf{GLM4}}     & \multicolumn{1}{c}{\textbf{LLaMA3}} & \multicolumn{1}{c}{\textbf{Reflection}} \\
        \multirow{-2}{*}{\textbf{}} & \multicolumn{1}{c}{\textbf{2.7B}}     & \multicolumn{1}{c}{\textbf{3.8B}} & \multicolumn{1}{c}{\textbf{3.8B}}         & \multicolumn{1}{c}{\textbf{3.8Bx16}}     & \multicolumn{1}{c}{\textbf{3.8Bx16}} & \multicolumn{1}{c}{\textbf{7B}}       & \multicolumn{1}{c}{\textbf{7B}}     & \multicolumn{1}{c}{\textbf{7B}}      & \multicolumn{1}{c}{\textbf{7B}}    & \multicolumn{1}{c}{\textbf{9B}}       & \multicolumn{1}{c}{\textbf{70B}}    & \multicolumn{1}{c}{\textbf{70B}}        \\ \midrule
        MMLU                        & {\color[HTML]{999999} 23.83}          & {\color[HTML]{999999} 67.27}      & {\color[HTML]{B7B7B7} 68.64}              & {\color[HTML]{B7B7B7} 76.62}             & {\color[HTML]{B7B7B7} 77.55}         & {\color[HTML]{999999} 47.08}          & {\color[HTML]{999999} 44.88}        & {\color[HTML]{999999} 57.29}         & {\color[HTML]{999999} 69.05}       & {\color[HTML]{999999} 67.36}          & {\color[HTML]{999999} 78.55}        & {\color[HTML]{999999} 75.83}            \\
        MMLU-g                      & {\color[HTML]{999999} 25.02}          & {\color[HTML]{999999} 85.29}      & {\color[HTML]{B7B7B7} 87}                 & {\color[HTML]{B7B7B7} 91.65}             & {\color[HTML]{B7B7B7} 92.83}         & {\color[HTML]{999999} 68.37}          & {\color[HTML]{999999} 72.87}        & {\color[HTML]{999999} 82.71}         & {\color[HTML]{999999} 89.23}       & {\color[HTML]{999999} 84.91}          & {\color[HTML]{999999} 92.17}        & {\color[HTML]{999999} 88.37}            \\
        \textit{difference}         & \cellcolor[HTML]{F3F3F3}\textbf{1.20} & \cellcolor[HTML]{F3F3F3}18.02     & \cellcolor[HTML]{F3F3F3}18.36             & \cellcolor[HTML]{F3F3F3}15.03            & \cellcolor[HTML]{F3F3F3}15.28        & \cellcolor[HTML]{F3F3F3}21.29         & \cellcolor[HTML]{F3F3F3}27.99       & \cellcolor[HTML]{F3F3F3}25.42        & \cellcolor[HTML]{F3F3F3}20.18      & \cellcolor[HTML]{F3F3F3}17.55         & \cellcolor[HTML]{F3F3F3}13.62       & \cellcolor[HTML]{F3F3F3}12.54           \\ \midrule
        ARC-C                       & {\color[HTML]{999999} 42.92}          & {\color[HTML]{999999} 80.20}      & {\color[HTML]{B7B7B7} 59.56}              & {\color[HTML]{B7B7B7} 54.78}             & {\color[HTML]{B7B7B7} 63.57}         & {\color[HTML]{999999} 50.34}          & {\color[HTML]{999999} 36.18}        & {\color[HTML]{999999} 64.08}         & {\color[HTML]{999999} 84.81}       & {\color[HTML]{999999} 86.35}          & {\color[HTML]{999999} 61.52}        & {\color[HTML]{999999} 56.74}            \\
        ARC-C-g                     & {\color[HTML]{999999} 47.27}          & {\color[HTML]{999999} 92.15}      & {\color[HTML]{B7B7B7} 93.94}              & {\color[HTML]{B7B7B7} 96.5}              & {\color[HTML]{B7B7B7} 96.25}         & {\color[HTML]{999999} 66.04}          & {\color[HTML]{999999} 44.71}        & {\color[HTML]{999999} 85.75}         & {\color[HTML]{999999} 95.22}       & {\color[HTML]{999999} 91.81}          & {\color[HTML]{999999} 95.99}        & {\color[HTML]{999999} 94.45}            \\
        \textit{difference}         & \cellcolor[HTML]{F3F3F3}\textbf{4.35} & \cellcolor[HTML]{F3F3F3}11.95     & \cellcolor[HTML]{F3F3F3}34.38             & \cellcolor[HTML]{F3F3F3}41.72            & \cellcolor[HTML]{F3F3F3}32.68        & \cellcolor[HTML]{F3F3F3}15.70         & \cellcolor[HTML]{F3F3F3}8.53        & \cellcolor[HTML]{F3F3F3}21.67        & \cellcolor[HTML]{F3F3F3}10.41      & \cellcolor[HTML]{F3F3F3}\textbf{5.46} & \cellcolor[HTML]{F3F3F3}34.47       & \cellcolor[HTML]{F3F3F3}37.71           \\ \midrule
        MathQA                      & {\color[HTML]{999999} 31.32}          & {\color[HTML]{999999} 41.14}      & {\color[HTML]{B7B7B7} 41.14}              & {\color[HTML]{B7B7B7} 37.42}             & {\color[HTML]{B7B7B7} 47.67}         & {\color[HTML]{999999} 34.30}          & {\color[HTML]{999999} 28.71}        & {\color[HTML]{999999} 36.88}         & {\color[HTML]{999999} 44.36}       & {\color[HTML]{999999} 43.05}          & {\color[HTML]{999999} 56.52}        & {\color[HTML]{999999} 58.29}            \\
        MathQA-g                    & {\color[HTML]{999999} 38.70}          & {\color[HTML]{999999} 49.06}      & {\color[HTML]{B7B7B7} 47.38}              & {\color[HTML]{B7B7B7} 43.96}             & {\color[HTML]{B7B7B7} 56.27}         & {\color[HTML]{999999} 35.71}          & {\color[HTML]{999999} 36.18}        & {\color[HTML]{999999} 45.77}         & {\color[HTML]{999999} 49.03}       & {\color[HTML]{999999} 56.04}          & {\color[HTML]{999999} 61.84}        & {\color[HTML]{999999} 63.92}            \\
        \textit{difference}         & \cellcolor[HTML]{F3F3F3}7.38          & \cellcolor[HTML]{F3F3F3}7.92      & \cellcolor[HTML]{F3F3F3}6.24              & \cellcolor[HTML]{F3F3F3}6.54             & \cellcolor[HTML]{F3F3F3}8.6          & \cellcolor[HTML]{F3F3F3}\textbf{1.41} & \cellcolor[HTML]{F3F3F3}7.47        & \cellcolor[HTML]{F3F3F3}8.89         & \cellcolor[HTML]{F3F3F3}4.67       & \cellcolor[HTML]{F3F3F3}12.99         & \cellcolor[HTML]{F3F3F3}5.32        & \cellcolor[HTML]{F3F3F3}5.63            \\ \bottomrule
        \end{tabular}
}
\end{adjustbox}
\caption{Detecting inadvertent contamination in popular open-sourced LLMs. Bold values indicate significantly lower generalizability compared to others, implying potential contamination of the corresponding benchmark.}
\label{tab:public}
\end{table*}

\section{Beyond Contamination}
\label{sec:discuss}
Can cross-lingual contamination only be utilized for cheating on benchmarks? In this section, we further discuss two potential scenarios where cross-lingual contamination can serve as a good starting point: interpreting the working mechanisms of LLMs~(\cref{sec:think}) and improving LLMs' unbalanced multilingual capabilities~(\cref{sec:local}).

\subsection{How Do LLMs Think Across Languages?}
\label{sec:think}
From~\cref{tab:cross_lingual}, we observe that the performance of the \textbf{same} backbone model can vary significantly when continually pre-trained on the \textbf{same} benchmark data in different languages. This is intriguing as we are injecting the \textbf{same} amount of knowledge.

Our hypothesis is that the knowledge in a model can be fixed, and language acts as an interface. Due to the uneven distribution of languages in the training corpus, the model's ability to understand and generate text can vary across different languages, which can be regarded as interfaces with varying qualities. In this case, despite the model having the same underlying knowledge, its performance can vary significantly, depending on the quality of the interfaces through which it is adopted.

\citet{wendler2024llamas} propose a similar idea that LLMs operate in ``input'', ``concept'', and ``output'' spaces when processing non-English. The input and output spaces here are similar to the language interfaces in our assumption. \citet{huang2024mindmerger} enhance LLMs' multilingual ability by feeding LLMs the encoded representation instead of the text of non-English inputs, which is also consistent with our hypothesis of language interfaces.

Therefore, we believe cross-lingual contamination can be a promising starting point for exploring the interpretability of multilingual LLMs.

\subsection{How to Localize LLMs for Non-English?}
\label{sec:local}
Considering a scenario where the budget is limited and we want a model with the best overall multilingual performance, in which single language should we conduct the continual pre-training?

As noted in~\cref{sec:inject_perform}, contamination in non-English languages can improve performance on the English benchmark. We further extend the evaluation to non-English languages to assess the impact of contamination on multilingual performance.

\begin{figure}[t]
    \centering
    \includegraphics[width=\linewidth]{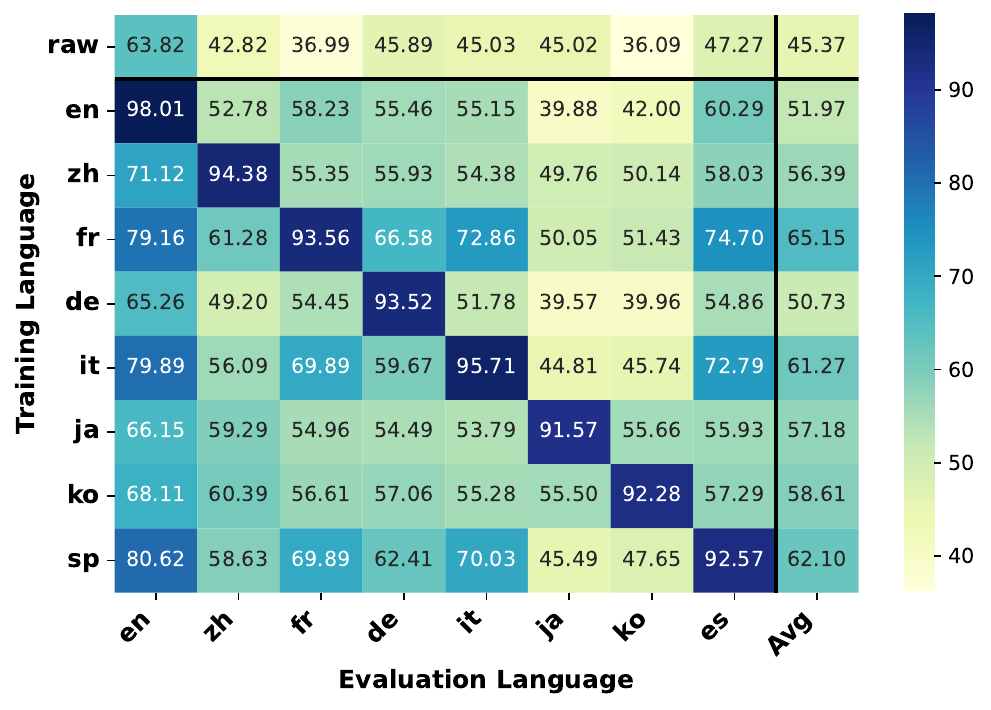}
    \caption{Performance (\%) of clean and contaminated (Y-axis) LLaMA3-8B on different language versions (X-axis) of MMLU. Here, the first row ``\textbf{raw}'' represents the clean model's performance. The rightmost column ``\textbf{Avg}'' shows the model's average performance across different language versions of MMLU. }
    \label{fig:heat}
\end{figure}

\cref{fig:heat} shows that contaminating in French achieves the best average performance, indicating that French could be the best choice for continual pre-training. Surprisingly, English only scored 51.97, ranking second last in all languages. 

Hence, investigating cross-lingual contamination can provide valuable perspectives for enhancing the unbalanced multilingual capabilities of LLMs.

\section{Related Work}
\subsection{Contamination Detection}
There has been a series of works 
for contamination detection. 
Mainly, they rely on a hypothesis that the test set is left in the training corpus in its original form.
Hence it is possible to detect contamination by examining the perplexity of the test set~\cite{jiang2024investigating}, or by asking the model to generate candidate choices and compare the similarity between the generated choice and original choice~\cite{golchin2023time}, or by checking if the order of questions/choices would have an impact on model performance~\cite{oren2023proving}.

However, these methods, while valuable, have certain limitations. The common assumption may not hold as simple paraphrasing can alter the training distribution, potentially evading the perplexity/n-gram check~\cite{jiang2024investigating}. Similarly, the wrong choices in multiple-choice benchmarks can be resampled and replaced to evade generation-style detection~\cite{golchin2023time}, and sequence order sensitivity~\cite{oren2023proving} can be alleviated via in-sample shuffling.


\subsection{Cross-Lingual Language Modeling}
Model's cross-lingual transferability has been extensively explored in recent years, particularly with the advent of Transformer models like BERT~\cite{devlin2018bert} and GPT2~\cite{radford2019language}. 
These models have been demonstrated to effectively leverage shared linguistic features across languages, enhancing their performance on cross-lingual tasks without the need for extensive language-specific training data. For instance, studies such as XLM-R~\cite{conneau-etal-2020-unsupervised}, which uses a transformer-based architecture to learn language-agnostic representations, show significant improvements in cross-lingual classification tasks. Similarly, \citet{wu2019beto} investigated the transferability of monolingual models to other languages by fine-tuning on small amounts of target language data, revealing that even limited adaptation can yield substantial gains in model performance across diverse language settings.


\section{Conclusions and Future Work}
In this paper, we identify a cross-lingual form of data contamination that can significantly inflate LLMs' benchmark performance while evading current detection approaches. In this case, we argue that the conventional memorization-based definition of data contamination could note reflect the essence of contamination. To detect such deeply concealed contamination, suggest a generalization-based definition and propose to detect it by examining the model's generalizability. With extensive experiments, we confirm that data contamination can cross language barriers. We also demonstrate that our proposed generalization-based method is able to detect not only cross-lingual but also other undisclosed contamination. In the future, we will extend our generalization-based detection approach to other potential forms of contamination. We will also explore how such cross-lingual contamination can benefit the interpretability of LLMs and the enhancement of multilingual capabilities.
\section*{Limitations}
Although we conducted extensive experiments on both the injection and detection of cross-lingual contamination, the investigation of this work has some limitations: \textbf{(1)} The injection of cross-lingual contamination is only based on 7B LLMs. Whether such cross-lingual contamination universally works on other sizes of LLMs is unclear. \textbf{(2)} The benchmarks we select are all multiple-choice questions-answering, which limits the detection of contamination on other forms of benchmarks. We select the multiple-choice datasets as they are among the most widely adopted benchmarks for LLMs evaluation. \textbf{(3)} The contamination for different benchmarks and languages is injected separately, which may not reflect the real-world scenarios where multiple benchmarks and languages are blended. The main reason for not including such a multilingual and multi-benchmark mixture is the constraint on computation resources, as we employ full-parameter continual pre-training instead of parameter-efficient fine-tuning. We encourage future works to tackle these limitations and provide stronger detection methods to uncover the potential undisclosed contamination in the wild.

\section*{Ethical Considerations}
We discuss the ethical considerations and broader impact of our work here: (1) \textbf{Intended Use.} We identify cross-lingual contamination to remind the community of the risk of such deeply concealed contamination. Our proposed detection method is to inspire future works to unmask other undisclosed contamination. (2) \textbf{Misuse Risks.} The experimental results and findings in this paper \textbf{should not} be used for offensive arguments or interpreted as implying misconduct of other works.

\bibliography{custom.bib}

\begin{thebibliography}{35}
\expandafter\ifx\csname natexlab\endcsname\relax\def\natexlab#1{#1}\fi

\bibitem[{Amini et~al.(2019)Amini, Gabriel, Lin, Koncel-Kedziorski, Choi, and Hajishirzi}]{amini-etal-2019-mathqa}
Aida Amini, Saadia Gabriel, Shanchuan Lin, Rik Koncel-Kedziorski, Yejin Choi, and Hannaneh Hajishirzi. 2019.
\newblock \href {https://doi.org/10.18653/v1/N19-1245} {{M}ath{QA}: Towards interpretable math word problem solving with operation-based formalisms}.
\newblock In \emph{Proceedings of the 2019 Conference of the North {A}merican Chapter of the Association for Computational Linguistics: Human Language Technologies, Volume 1 (Long and Short Papers)}, pages 2357--2367, Minneapolis, Minnesota. Association for Computational Linguistics.

\bibitem[{Bai and et~al(2023)}]{qwen}
Jinze Bai and et~al. 2023.
\newblock Qwen technical report.
\newblock \emph{arXiv preprint arXiv:2309.16609}.

\bibitem[{Beeching et~al.(2023)Beeching, Fourrier, Habib, Han, Lambert, Rajani, Sanseviero, Tunstall, and Wolf}]{open-llm-leaderboard}
Edward Beeching, Clémentine Fourrier, Nathan Habib, Sheon Han, Nathan Lambert, Nazneen Rajani, Omar Sanseviero, Lewis Tunstall, and Thomas Wolf. 2023.
\newblock Open llm leaderboard.
\newblock \url{https://huggingface.co/spaces/open-llm-leaderboard/open_llm_leaderboard}.

\bibitem[{Brown et~al.(2020)Brown, Mann, Ryder, Subbiah, Kaplan, Dhariwal, Neelakantan, Shyam, Sastry, Askell et~al.}]{brown2020language}
Tom Brown, Benjamin Mann, Nick Ryder, Melanie Subbiah, Jared~D Kaplan, Prafulla Dhariwal, Arvind Neelakantan, Pranav Shyam, Girish Sastry, Amanda Askell, et~al. 2020.
\newblock \href {https://papers.nips.cc/paper/2020/file/1457c0d6bfcb4967418bfb8ac142f64a-Paper.pdf} {Language models are few-shot learners}.
\newblock In \emph{Proceedings of NeurIPS}, volume~33, pages 1877--1901.

\bibitem[{Chern et~al.(2023)Chern, Zou, Li, Hu, Feng, Li, and Liu}]{abel}
Ethan Chern, Haoyang Zou, Xuefeng Li, Jiewen Hu, Kehua Feng, Junlong Li, and Pengfei Liu. 2023.
\newblock Generative ai for math: Abel.
\newblock \url{https://github.com/GAIR-NLP/abel}.

\bibitem[{Chowdhery et~al.(2023)Chowdhery, Narang, Devlin, Bosma, Mishra, Roberts, Barham, Chung, Sutton, Gehrmann et~al.}]{chowdhery2023palm}
Aakanksha Chowdhery, Sharan Narang, Jacob Devlin, Maarten Bosma, Gaurav Mishra, Adam Roberts, Paul Barham, Hyung~Won Chung, Charles Sutton, Sebastian Gehrmann, et~al. 2023.
\newblock Palm: Scaling language modeling with pathways.
\newblock \emph{Journal of Machine Learning Research}, 24(240):1--113.

\bibitem[{Clark et~al.(2018)Clark, Cowhey, Etzioni, Khot, Sabharwal, Schoenick, and Tafjord}]{clark2018think}
Peter Clark, Isaac Cowhey, Oren Etzioni, Tushar Khot, Ashish Sabharwal, Carissa Schoenick, and Oyvind Tafjord. 2018.
\newblock Think you have solved question answering? try arc, the ai2 reasoning challenge.
\newblock \emph{arXiv preprint arXiv:1803.05457}.

\bibitem[{Conneau and et~al(2020)}]{conneau-etal-2020-unsupervised}
Alexis Conneau and et~al. 2020.
\newblock \href {https://doi.org/10.18653/v1/2020.acl-main.747} {Unsupervised cross-lingual representation learning at scale}.
\newblock In \emph{Proceedings of the 58th Annual Meeting of the Association for Computational Linguistics}, pages 8440--8451, Online. Association for Computational Linguistics.

\bibitem[{Deng et~al.(2023)Deng, Zhao, Tang, Gerstein, and Cohan}]{deng2023investigating}
Chunyuan Deng, Yilun Zhao, Xiangru Tang, Mark Gerstein, and Arman Cohan. 2023.
\newblock Investigating data contamination in modern benchmarks for large language models.
\newblock \emph{arXiv preprint arXiv:2311.09783}.

\bibitem[{Devlin et~al.(2018)Devlin, Chang, Lee, and Toutanova}]{devlin2018bert}
Jacob Devlin, Ming-Wei Chang, Kenton Lee, and Kristina Toutanova. 2018.
\newblock Bert: Pre-training of deep bidirectional transformers for language understanding.
\newblock \emph{arXiv preprint arXiv:1810.04805}.

\bibitem[{Dodge et~al.(2021)Dodge, Sap, Marasovi{\'c}, Agnew, Ilharco, Groeneveld, Mitchell, and Gardner}]{dodge2021documenting}
Jesse Dodge, Maarten Sap, Ana Marasovi{\'c}, William Agnew, Gabriel Ilharco, Dirk Groeneveld, Margaret Mitchell, and Matt Gardner. 2021.
\newblock Documenting large webtext corpora: A case study on the colossal clean crawled corpus.
\newblock \emph{arXiv preprint arXiv:2104.08758}.

\bibitem[{Dong et~al.(2024)Dong, Jiang, Liu, Jin, and Li}]{dong2024generalization}
Yihong Dong, Xue Jiang, Huanyu Liu, Zhi Jin, and Ge~Li. 2024.
\newblock Generalization or memorization: Data contamination and trustworthy evaluation for large language models.
\newblock \emph{arXiv preprint arXiv:2402.15938}.

\bibitem[{Gao et~al.(2023)Gao,  et~al.}]{eval-harness}
Leo Gao, , et~al. 2023.
\newblock \href {https://doi.org/10.5281/zenodo.10256836} {A framework for few-shot language model evaluation}.

\bibitem[{Golchin and Surdeanu(2023)}]{golchin2023time}
Shahriar Golchin and Mihai Surdeanu. 2023.
\newblock Time travel in llms: Tracing data contamination in large language models.
\newblock In \emph{The Twelfth International Conference on Learning Representations}.

\bibitem[{Gunasekar et~al.(2023)Gunasekar, Zhang, Aneja, Mendes, Del~Giorno, Gopi, Javaheripi, Kauffmann, de~Rosa, Saarikivi et~al.}]{gunasekar2023textbooks}
Suriya Gunasekar, Yi~Zhang, Jyoti Aneja, Caio C{\'e}sar~Teodoro Mendes, Allie Del~Giorno, Sivakanth Gopi, Mojan Javaheripi, Piero Kauffmann, Gustavo de~Rosa, Olli Saarikivi, et~al. 2023.
\newblock Textbooks are all you need.
\newblock \emph{arXiv preprint arXiv:2306.11644}.

\bibitem[{Hendrycks et~al.(2020)Hendrycks, Burns, Basart, Zou, Mazeika, Song, and Steinhardt}]{hendrycks2020measuring}
Dan Hendrycks, Collin Burns, Steven Basart, Andy Zou, Mantas Mazeika, Dawn Song, and Jacob Steinhardt. 2020.
\newblock Measuring massive multitask language understanding.
\newblock In \emph{International Conference on Learning Representations}.

\bibitem[{Huang et~al.(2024)Huang, Zhu, Cheng, Li, and Yuan}]{huang2024mindmerger}
Zixian Huang, Wenhao Zhu, Gong Cheng, Lei Li, and Fei Yuan. 2024.
\newblock Mindmerger: Efficient boosting llm reasoning in non-english languages.
\newblock \emph{arXiv preprint arXiv:2405.17386}.

\bibitem[{Jiang et~al.(2024{\natexlab{a}})Jiang, Sablayrolles, Roux, Mensch, Savary, Bamford, Chaplot, Casas, Hanna, Bressand et~al.}]{jiang2024mixtral}
Albert~Q Jiang, Alexandre Sablayrolles, Antoine Roux, Arthur Mensch, Blanche Savary, Chris Bamford, Devendra~Singh Chaplot, Diego de~las Casas, Emma~Bou Hanna, Florian Bressand, et~al. 2024{\natexlab{a}}.
\newblock Mixtral of experts.
\newblock \emph{arXiv preprint arXiv:2401.04088}.

\bibitem[{Jiang et~al.(2024{\natexlab{b}})Jiang, Liu, Zhong, Schaeffer, Ouyang, Han, and Koyejo}]{jiang2024investigating}
Minhao Jiang, Ken~Ziyu Liu, Ming Zhong, Rylan Schaeffer, Siru Ouyang, Jiawei Han, and Sanmi Koyejo. 2024{\natexlab{b}}.
\newblock Investigating data contamination for pre-training language models.
\newblock \emph{arXiv preprint arXiv:2401.06059}.

\bibitem[{Lee et~al.(2023)Lee, Hunter, and Ruiz}]{lee2023platypus}
Ariel~N Lee, Cole~J Hunter, and Nataniel Ruiz. 2023.
\newblock Platypus: Quick, cheap, and powerful refinement of llms.
\newblock \emph{arXiv preprint arXiv:2308.07317}.

\bibitem[{Li(2023{\natexlab{a}})}]{li2023estimating}
Yucheng Li. 2023{\natexlab{a}}.
\newblock Estimating contamination via perplexity: Quantifying memorisation in language model evaluation.
\newblock \emph{arXiv preprint arXiv:2309.10677}.

\bibitem[{Li(2023{\natexlab{b}})}]{li2023open}
Yucheng Li. 2023{\natexlab{b}}.
\newblock An open source data contamination report for llama series models.
\newblock \emph{arXiv preprint arXiv:2310.17589}.

\bibitem[{Liu et~al.(2024)Liu, Kim, Wang, Liang, Shen, Cheng, Liu, Tanaka, Wu, Hu, Chaudhary, Lin, Zhang, Xue, Awadalla, Gao, and Chen}]{liu2024gringradientinformedmoe}
Liyuan Liu, Young~Jin Kim, Shuohang Wang, Chen Liang, Yelong Shen, Hao Cheng, Xiaodong Liu, Masahiro Tanaka, Xiaoxia Wu, Wenxiang Hu, Vishrav Chaudhary, Zeqi Lin, Chenruidong Zhang, Jilong Xue, Hany Awadalla, Jianfeng Gao, and Weizhu Chen. 2024.
\newblock \href {http://arxiv.org/abs/2409.12136} {Grin: Gradient-informed moe}.

\bibitem[{Meta(2024)}]{llama3}
Meta. 2024.
\newblock Introducing meta llama 3: The most capable openly available llm to date.

\bibitem[{Nasr et~al.(2023)Nasr, Carlini, Hayase, Jagielski, Cooper, Ippolito, Choquette-Choo, Wallace, Tram{\`e}r, and Lee}]{nasr2023scalable}
Milad Nasr, Nicholas Carlini, Jonathan Hayase, Matthew Jagielski, A~Feder Cooper, Daphne Ippolito, Christopher~A Choquette-Choo, Eric Wallace, Florian Tram{\`e}r, and Katherine Lee. 2023.
\newblock Scalable extraction of training data from (production) language models.
\newblock \emph{arXiv preprint arXiv:2311.17035}.

\bibitem[{OpenAI(2023)}]{openai2023gpt4}
OpenAI. 2023.
\newblock \href {http://arxiv.org/abs/2303.08774} {Gpt-4 technical report}.

\bibitem[{OpenAI(2024)}]{openai2024gpt4o}
OpenAI. 2024.
\newblock Hello gpt4o.

\bibitem[{Oren et~al.(2023)Oren, Meister, Chatterji, Ladhak, and Hashimoto}]{oren2023proving}
Yonatan Oren, Nicole Meister, Niladri~S Chatterji, Faisal Ladhak, and Tatsunori Hashimoto. 2023.
\newblock Proving test set contamination for black-box language models.
\newblock In \emph{The Twelfth International Conference on Learning Representations}.

\bibitem[{Radford et~al.(2019)Radford, Wu, Child, Luan, Amodei, Sutskever et~al.}]{radford2019language}
Alec Radford, Jeffrey Wu, Rewon Child, David Luan, Dario Amodei, Ilya Sutskever, et~al. 2019.
\newblock Language models are unsupervised multitask learners.
\newblock \emph{OpenAI blog}, 1(8):9.

\bibitem[{Riddell et~al.(2024)Riddell, Ni, and Cohan}]{riddell2024quantifying}
Martin Riddell, Ansong Ni, and Arman Cohan. 2024.
\newblock Quantifying contamination in evaluating code generation capabilities of language models.
\newblock \emph{arXiv preprint arXiv:2403.04811}.

\bibitem[{Shi et~al.(2023)Shi, Ajith, Xia, Huang, Liu, Blevins, Chen, and Zettlemoyer}]{shi2023detecting}
Weijia Shi, Anirudh Ajith, Mengzhou Xia, Yangsibo Huang, Daogao Liu, Terra Blevins, Danqi Chen, and Luke Zettlemoyer. 2023.
\newblock Detecting pretraining data from large language models.
\newblock \emph{arXiv preprint arXiv:2310.16789}.

\bibitem[{Touvron et~al.(2023)Touvron, Martin, Stone, Albert, Almahairi, Babaei, Bashlykov, Batra, Bhargava, Bhosale et~al.}]{touvron2023llama}
Hugo Touvron, Louis Martin, Kevin Stone, Peter Albert, Amjad Almahairi, Yasmine Babaei, Nikolay Bashlykov, Soumya Batra, Prajjwal Bhargava, Shruti Bhosale, et~al. 2023.
\newblock \href {https://arxiv.org/abs/2307.09288} {Llama 2: Open foundation and fine-tuned chat models}.
\newblock \emph{ArXiv preprint}, abs/2307.09288.

\bibitem[{Wendler et~al.(2024)Wendler, Veselovsky, Monea, and West}]{wendler2024llamas}
Chris Wendler, Veniamin Veselovsky, Giovanni Monea, and Robert West. 2024.
\newblock Do llamas work in english? on the latent language of multilingual transformers.
\newblock \emph{arXiv preprint arXiv:2402.10588}.

\bibitem[{Wu and Dredze(2019)}]{wu2019beto}
Shijie Wu and Mark Dredze. 2019.
\newblock Beto, bentz, becas: The surprising cross-lingual effectiveness of bert.
\newblock \emph{arXiv preprint arXiv:1904.09077}.

\bibitem[{Xu et~al.(2024)Xu, Wang, Fan, and Liu}]{xu2024benchmarking}
Ruijie Xu, Zengzhi Wang, Run-Ze Fan, and Pengfei Liu. 2024.
\newblock Benchmarking benchmark leakage in large language models.
\newblock \emph{arXiv preprint arXiv:2404.18824}.

\end{thebibliography}

\appendix
\clearpage
\section*{Appendices}
\section{Details for Contamination Injection}
In the experiments of injecting cross-lingual contamination, we adopt three widely adopted public benchmarks and translate their test sets into different languages for continual pre-training on two open-sourced multilingual LLMs.

\subsection{Benchmark Test Sets}
The benchmark datasets we use are all in the form of multiple-choice, which are licensed and intended for research use. Their details are as follows.


\textbf{MMLU}\footnote{\url{https://huggingface.co/datasets/hails/mmlu_no_train}}\cite{hendrycks2020measuring} is a benchmark for measuring models' language understanding ability with questions in various domains, such as biology, engineering, and computer science. The test set contains around $14k$ questions in total.

\textbf{ARC-Challenge}\footnote{\url{https://huggingface.co/datasets/allenai/ai2_arc}}\cite{clark2018think} is a dataset specially designed for the evaluation of reasoning ability. Its test set consists of $2.59k$ data samples.

\textbf{MathQA}\footnote{\url{https://huggingface.co/datasets/allenai/math_qa}}\cite{amini-etal-2019-mathqa} is a professional mathematical question-answering dataset of which the choices are mostly Arabic numbers. There are around $2.99k$ questions in the test set.

\subsection{Translation Prompt}
The quality of translation is critical for our experiments. Therefore, considering both cost and quality, we utilized LLaMA3\footnote{\url{https://huggingface.co/meta-llama/Meta-Llama-3-8B-instruct}} to conduct the translations. The prompt template is shown below.
\label{sec:tran_tmp}
\begin{lstlisting}[language=Python, numbers=none, label={lst:tran_tmp}]
"Help me translate the following text into native <language>: <text>. do not use direct translation. Output your translation only without any explanations or notes! Output your translation only without any explanations or notes! Output your translation only without any explanations or notes!"
\end{lstlisting}

\subsection{Continual Pre-Training}
We employ continual pre-training to contaminate two multilingual LLMs (LLaMA3-8B and Qwen1.5-7B) with the original English and translated versions of benchmark test sets. 
The training hyperparameters are shown in~\cref{tab:hyper_parameters}. The experiment is conducted on Nvidia Tesla A100 GPUs.

\begin{table}[h]
\centering
\begin{adjustbox}{max width=0.7\linewidth}{
\begin{tabular}{lc}
\toprule
Batch Size    & $16$             \\
Learning Rate & $5\times10^{-5}$           \\
Optimizer     & AdaFactor      \\
Epochs        & $36$    \\
\bottomrule
\end{tabular}}
\end{adjustbox}
\caption{Hyperparameters for continual pre-training}
\label{tab:hyper_parameters}
\end{table}

\section{Details for Contamination Detection}
For contamination detection, we implement three baselines along with our proposed generalization-based method (choice confusion). The experiments of contamination detection are conducted on Nvidia RTX886 A6000 GPUs.

\subsection{Shared Likelihood}
Our implementation is largely based on the original codebase\footnote{\url{https://github.com/tatsu-lab/test_set_contamination}} provided by \citet{golchin2023time}. To ensure a fair evaluation, we first try to reproduce the results in \citet{golchin2023time} and then adapt the code to our scenario. Due to the randomness of the permutation test and the selection of parameters in the original implementation, our reproduced results are slightly different than those in the paper but consistent in general.

\subsection{Guided Prompting}
\label{sec:judge_tmp}

We adopt GPT-4o~\cite{openai2024gpt4o} with in-context examples to judge if the model's predicted choice essentially has the same meaning as the correct one. The specific prompt template is shown below.

\begin{lstlisting}[language=Python, numbers=none]
"<question>
Compare the following two sentences and determine if they have the same meaning. Answer with "true" if they do and "false" if they do not. No Explanation needed, do not repeat question.

Example1:
<example1>
Sentence 1: The sky is blue.
Sentence 2: The sky is clear.
Answer: false
</example1>

Example2:
<example2>
Sentence 1: She is a doctor.
Sentence 2: She practices medicine.
Answer: true
</example2>

Now, compare these sentences:

<class>
Sentence 1: [{i[0]}]
Sentence 2: [{i[1]}]
        
Do the two sentences have the same meaning? Answer with "true" if they do and "false" if they do not
Your Answer:
</class>
</question>"
\end{lstlisting}

\subsection{N-Gram Accuracy}

We adopt a similar approach to that used by \citet{xu2024benchmarking}. Instead of calculating the n-gram accuracy on the combined text of the question and answer, we focus on the question and choices. We identify five equally spaced indices within the combined tokens. For each index, we provide the model with the prefix text preceding the index and then determine the n-gram accuracy of the generated text. The n-gram accuracy is expected to be higher if the model is contaminated, as then the generated tokens will be more similar to the tokens within the dataset. The pseudocode for the n-gram accuracy calculation process is shown as follows. 

\begin{lstlisting}[language=Python, numbers=none,label={lst:n_gram}]
# Create combined question and choice text
format_text = f"{question}{choice}"
tokens = tokenizer.tokenize(format_text)
# Find indexes for prefix texts
starting_points = np.linspace(2, len(tokens), num=5)

correct_n_grams = 0
total_n_grams = 0
for idx in starting_points:
    # Generate text based on prefix text
    gens = model.generate(tokens[:idx])
    total_n_grams += 1
    # Compare generated and original n gram tokens
    if gens[0, -n:] == tokens[idx:idx + n]):
        correct_n_grams += 1
# Calculate n-gram accuracy
n_gram_accuracy = correct_n_grams / total_n_grams
\end{lstlisting}
\label{sec:ngram}

\subsection{Choice Confusion}
\label{app:choice_confusion}

We utilize the LM-Eval\footnote{\url{https://github.com/EleutherAI/lm-evaluation-harness}} framework to evaluate different models on the original and translated versions of benchmarks to ensure fair comparisons. 

The experiments of contamination detection are not limited to detecting the cross-lingual contamination injected by us intentionally. We also detect other undisclosed contamination in real-world popular multi-lingual LLMs, including Phi2-2.7B\footnote{\url{https://huggingface.co/microsoft/phi-2}}, Phi3-3.8B\footnote{\url{https://huggingface.co/microsoft/Phi-3-mini-4k-instruct}}, Phi3.5-mini\footnote{\url{https://huggingface.co/microsoft/Phi-3.5-mini-instruct}}, Phi3.5-MoE\footnote{\url{https://huggingface.co/microsoft/Phi-3.5-MoE-instruct}}, GrinMoE\footnote{\url{https://huggingface.co/microsoft/GRIN-MoE}}~\cite{liu2024gringradientinformedmoe}, Abel-7B\footnote{\url{https://huggingface.co/GAIR/Abel-7B-002}}~\cite{abel}, LLaMA2-7B\footnote{\url{https://huggingface.co/meta-llama/Llama-2-7b-chat-hf}}, Mistral-7B\footnote{\url{https://huggingface.co/mistralai/Mistral-7B-Instruct-v0.2}}, GLM4-9B\footnote{\url{https://huggingface.co/THUDM/glm-4-9b-chat}}, Qwen2-7B\footnote{\url{https://huggingface.co/Qwen/Qwen2-7B-Instruct}}, LLaMA3-70b\footnote{\url{https://huggingface.co/meta-llama/Meta-Llama-3-70B}}, Reflection-70B\footnote{\url{https://huggingface.co/mattshumer/Reflection-Llama-3.1-70B}}.

In the LM-Eval framework, the specific yaml templates we use for MMLU, ARC-Challenge, and MathQA  are provided as follows.
\vspace{1em}

\begin{lstlisting}[language=Python, numbers=none, label={lst:mmlu_temp}]
# MMLU Template
task: custom_mmlu_name
dataset_path: custom_mmlu_datapath
test_split: test
fewshot_config:
  sampler: first_n
output_type: multiple_choice
doc_to_text: "{{question.strip()}}\nA. {{choices[0]}}\nB. {{choices[1]}}\nC. {{choices[2]}}\nD. {{choices[3]}}\nAnswer:"
doc_to_choice: ["A", "B", "C", "D"]
doc_to_target: answer
metric_list:
  - metric: acc
    aggregation: mean
    higher_is_better: true
metadata:
  version: 0.0
\end{lstlisting}

\begin{lstlisting}[language=Python, numbers=none,label={lst:arc_temp}]
# ARC-Challenge Template
group:
  - ai2_arc
task: custom_arc_name
dataset_path: custom_arc_datapath
output_type: multiple_choice
test_split: test
doc_to_text: "Question: {{question}}\nChoices: {{choices.text}}\nOptions:{{choices.label}}\nAnswer:"
doc_to_choice: "{{choices.label}}"
doc_to_target: "{{choices.label.index(answerKey)}}"
should_decontaminate: true
doc_to_decontamination_query: "Question: {{question}}\nAnswer:"
metric_list:
  - metric: acc
    aggregation: mean
    higher_is_better: true
  - metric: acc_norm
    aggregation: mean
    higher_is_better: true
metadata:
  version: 1.0
\end{lstlisting}

\begin{lstlisting}[language=Python, numbers=none,label={lst:mathqa_temp}]
#MathQA Template
task: custom_mathqa_name
dataset_path: custom_mathqa_datapath
output_type: multiple_choice
test_split: test
doc_to_text: "Question: {{Problem}}\nAnswer:"
doc_to_target: "{{['a', 'b', 'c', 'd', 'e'].index(correct)}}"
doc_to_choice: !function utils.doc_to_choice
should_decontaminate: true
doc_to_decontamination_query: "Question: {{Problem}}\nAnswer:"
metric_list:
  - metric: acc
    aggregation: mean
    higher_is_better: true
  - metric: acc_norm
    aggregation: mean
    higher_is_better: true
metadata:
  version: 1.0
\end{lstlisting}

There are mainly 5 hyperparameters: \texttt{Model Path}, \texttt{Task}, \texttt{Batch Size}, \texttt{Max Batch Size}, \texttt{N shot}. \texttt{Model Path} and \texttt{Task} will be set as custom paths and names, and we set \texttt{Batch Size} and \texttt{Max Batch Size} to 2 and \texttt{N shot} as 0.

\end{document}